\documentclass[conference]{IEEEtran}
\usepackage{algorithm}
\pdfoutput=1
\usepackage{algorithmic}
\usepackage{cite}
\usepackage{amsmath,amssymb,amsfonts}
\usepackage{graphicx}
\usepackage{textcomp}
\usepackage{xcolor}
\def\BibTeX{{\rm B\kern-.05em{\sc i\kern-.025em b}\kern-.08em
    T\kern-.1667em\lower.7ex\hbox{E}\kern-.125emX}}
\usepackage{threeparttable}
\usepackage{microtype}
\usepackage{graphicx}
\usepackage{booktabs}
\usepackage{hyperref}
\usepackage{caption}
\usepackage{catchfilebetweentags}
\usepackage{url}
\usepackage{xspace}
\usepackage{comment}
\usepackage[export]{adjustbox}
\usepackage{paralist}
\usepackage{enumitem}
\usepackage{catchfilebetweentags}
\usepackage{etoolbox}
\usepackage{adjustbox}
\usepackage{amssymb}
\usepackage[normalem]{ulem}
\usepackage{tcolorbox}

\usepackage{adjustbox}
\usepackage{booktabs}
\usepackage{multirow,mathtools } 
\usepackage{arydshln}
\usepackage{fontawesome}
\usepackage{pifont}
\usepackage{xcolor}
\definecolor{gblue}{rgb}{0.784, 0.851, 0.902}

\newcommand{\etc}{\textit{etc}.\xspace}

\newcommand{\ie}{\textit{i.e.}\xspace}
\newcommand{\eg}{\textit{e.g.}\xspace}
\newcommand{\etal}{\textit{et. al.}}

\newcommand{\subsec}[1]{\par\noindent\textbf{#1}}

\newcommand{\pf}{PF}
\newcommand{\ff}{FF}

\newcommand{\seqtseq}{\textsc{seq2seq}}

\newcommand{\replace}{\text{replace}}
\newcommand{\insertt}{\text{insert}}
\newcommand{\sites}{\textit{sites}}
\newcommand{\site}{\textit{site}}

\newcommand{\advtr}{\textsc{AT}}
\newcommand{\stdtr}{\textsc{ST}}

\newcommand{\modelnum}[1]{{{\fontfamily{qpl}\selectfont M}\textsubscript{#1}}}
\newcommand{\notate}[3]{{{#1}-{#3}}}

\newcommand{\genf}{\textsc{Gen-F1}}
\newcommand{\robf}{\textsc{Rob-F1}}

\newcommand{\prog}{\mathcal{P}}
\newcommand{\pnormal}{$\mathcal{P}$}
\newcommand{\prandom}{$\mathcal{P}_{\text{random}}$}
\newcommand{\padv}{$\mathcal{P}_{\text{adv}}$}

\newcommand{\ourmodel}{\textsc{ClawSAT}}
\newcommand{\claw}{\textsc{Claw}}
\newcommand{\sat}{\textsc{SAT}}
\newcommand{\cl}{Contrastive learning}
\newcommand{\cllower}{contrastive learning}
\newcommand{\clsmall}{CL}

\newcommand{\spc}[1]{#1\xspace}
\newcommand{\shash}[1]{\textcolor{red}{#1}}

\newcommand{\contracode}{\textsc{ContraCode}}
\newcommand{\summarypy}{\textsc{SummaryPy}}
\newcommand{\summaryjava}{\textsc{SummaryJava}}
\newcommand{\completepy}{\textsc{CompletePy}}

\newcommand{\clonejava}{\textsc{CloneJava}}

\newcommand{\pysmall}{\textsc{Py150}}
\newcommand{\pylarge}{\textsc{Py-CSN}}
\newcommand{\javasmall}{\textsc{Java-C2S}}
\newcommand{\javalarge}{\textsc{Java-CSN}}

\newcommand{\loadFig}[1]{
	\ExecuteMetaData[8figures_tables.tex]{#1}
}
\usepackage{amsmath}
\usepackage{amssymb}
\usepackage{mathtools}
\usepackage{amsthm}

\usepackage[capitalize,noabbrev]{cleveref}

\theoremstyle{plain}

\theoremstyle{definition}

\theoremstyle{remark}

\DeclareMathOperator*{\minimize}{\text{minimize}}

\DeclareMathOperator*{\st}{\text{subject to}}
\DeclareMathAlphabet\mathbfcal{OMS}{cmsy}{b}{n}
\newcommand{\Def}[0]{\mathrel{\mathop:}=}

\newcommand\figstrut[2]{
  \dimen0=#1%
  \advance\dimen0 by -#2%
  \divide\dimen0 by -2%
  \dimen1=#1%
  \advance\dimen1 by \dimen0%
  \vrule height \dimen1 depth \dimen0 width 0pt\relax%
}
\setlist[itemize]{leftmargin=*, nosep, align=parleft, noitemsep, topsep=0pt}

\usepackage{newfloat}
\usepackage{listings, lstautogobble}

\DeclareCaptionStyle{ruled}{labelfont=normalfont,labelsep=colon,strut=off} 
\def\BibTeX{{\rm B\kern-.05em{\sc i\kern-.025em b}\kern-.08em
    T\kern-.1667em\lower.7ex\hbox{E}\kern-.125emX}}
\begin{document}

\title{\ourmodel: Towards Both Robust and Accurate Code Models
}
\author{
\IEEEauthorblockN{Jinghan Jia\textsuperscript{1*}, Shashank Srikant\textsuperscript{2,3*}, Tamara Mitrovska\textsuperscript{2},  Chuang Gan\textsuperscript{3}, Shiyu Chang\textsuperscript{4}\\
Sijia Liu\textsuperscript{1,3}, Una-May {O'R}eilly\textsuperscript{2,3}\\~\\
${}^1$Michigan State University \quad ${}^2$CSAIL, MIT \quad ${}^3$MIT-IBM Watson AI Lab \quad ${}^4$UC Santa Barbara\\
*Equal contribution\\
Correspondence: jiajingh@msu.edu \quad shash@mit.edu 
}
}

\maketitle

\begin{abstract}
We  {integrate contrastive learning (CL) with adversarial learning to}
co-optimize   the robustness and accuracy of code models. 
{Different from existing works, we show that code obfuscation, a {standard} code transformation operation, provides novel means to generate complementary `views' of a code that enable us to achieve both robust and accurate code models.}
{To the best of our knowledge, this is the first systematic study to explore and exploit the robustness and accuracy benefits of (multi-view) code obfuscations in code models.}
Specifically, we first adopt adversarial codes as robustness-promoting views in CL at the self-supervised pre-training phase. This yields improved robustness and transferability for downstream tasks. 
Next, at the supervised fine-tuning stage, we show that adversarial training with a proper temporally-staggered schedule of adversarial code generation can further improve robustness and accuracy of the pre-trained code model.
Built on the above two modules, we develop \ourmodel, a novel self-supervised learning (SSL) framework for code by integrating \underline{CL} with \underline{a}dversarial  vie\underline{w}s ({\claw}) with \underline{s}taggered \underline{a}dversarial \underline{t}raining ({\sat}).
On evaluating three downstream tasks across Python and Java, we show that \spc\ourmodel consistently yields the best robustness and accuracy (\eg 11\% in robustness and 6\% in accuracy on the code summarization task in Python).
We additionally demonstrate the effectiveness of adversarial learning in {\claw} by analyzing the characteristics of the loss landscape and interpretability of the pre-trained models. Codes are available at https://github.com/OPTML-Group/CLAW-SAT.
\end{abstract}

\begin{IEEEkeywords}
Robustness, Programming languages, Deep Learning
\end{IEEEkeywords}
\section{Introduction}
\label{sec:introduction}

Recent progress in large language models for \textit{computer programs} (\ie code) suggests a growing interest in {self-supervised learning (\textbf{SSL})} methods to learn code models--deep learning models that process and reason about code \cite{kanade2020learning, chen2021evaluating, chen2021plur, jigsaw2022, jain-etal-2021-contrastive}. 
In these models, a task-agnostic encoder is learned in a pre-training step, typically on an unlabeled corpus. 
The encoder is appended to another predictive model which is then fine-tuned for a specific downstream task.
In particular, contrastive learning (\textbf{CL}) based self-supervision \cite{chen2020simple,he2020momentum} has shown to improve the downstream performance of code reasoning tasks when compared to state-of-the-art task-specific supervised learning (\textbf{SL}) models  \cite{cl1, cl2, jain-etal-2021-contrastive}.

While CL offers to be a promising SSL approach, {nearly all the existing works focus on improving the accuracy of code models. 
Yet, some very recent works \cite{ramakrishnan2020, yefet2020adversarial, srikant2021generating}} showed that trained supervised 
code models are vulnerable to \textbf{code obfuscation} transformations. 
These works propose \textbf{adversarial code}--changing a given code via obfuscation transformations. 
Such transformed programs retain the functionality of the original program but can fool a trained model at test time. {{Figure\,\ref{fig: adv_code_obf_examples}} shows an example of adversarial code achieved by code obfuscation (more details in Section\,\ref{sec:motivation}).}
{In software engineering, code obfuscation is}
a commonly-used method to hide code in software projects without altering their functionality \cite{collberg2002watermarking,
linn2003obfuscation}, and is consequently a popular choice among malware composers \cite{schrittwieser2016protecting}. 
Thus, it is important to study how obfuscation-based adversarial code could affect code model representations learned by SSL.
As an example, Schuster et al. \cite{schuster2021you} successfully demonstrate adversarial code attacks on a public code completion model pre-trained on GPT-2, a large language model of code.

Improving the robustness of ML models to adversarial code however comes at a cost--its accuracy (model generalization).
Works in vision \cite{Goodfellow2015explaining,madry2018towards}
and text \cite{miyato2016adversarial} have shown how adversarially trained models improve robustness at the cost of model accuracy.
While some works \cite{su2018robustness,tsipras2018robustness} provide a theoretical framework for how the robustness of learned models is always at odds with its accuracy, {this argument is mainly confined to the SL paradigm and vision applications, and thus remains uninvestigated in SSL for code.}
While there exist similarities between SSL in vision and code, the discrete and structured nature of inputs, and the additional constraints enforced on views (obfuscated codes) introduce a new set of challenges that have been unexplored by the vision community.
\begin{tcolorbox}[
	standard jigsaw,
	opacityback=0,  
	]
For code models, we ask: 
\textit{Can pre-trained models be made robust to adversarial attacks? 
Is it possible to retain this `pre-trained robustness' when fine-tuning on different tasks? 
And importantly, is it possible to improve on both the retained generalization and robustness during fine-tuning, thus challenging the popular view of having to trade-off robustness for accuracy gains? These questions form the focus of our work.}
\end{tcolorbox}
\subsection{Overview of proposed approach}

\begin{figure}[t]
    \centering

    \includegraphics[width=0.45\textwidth]{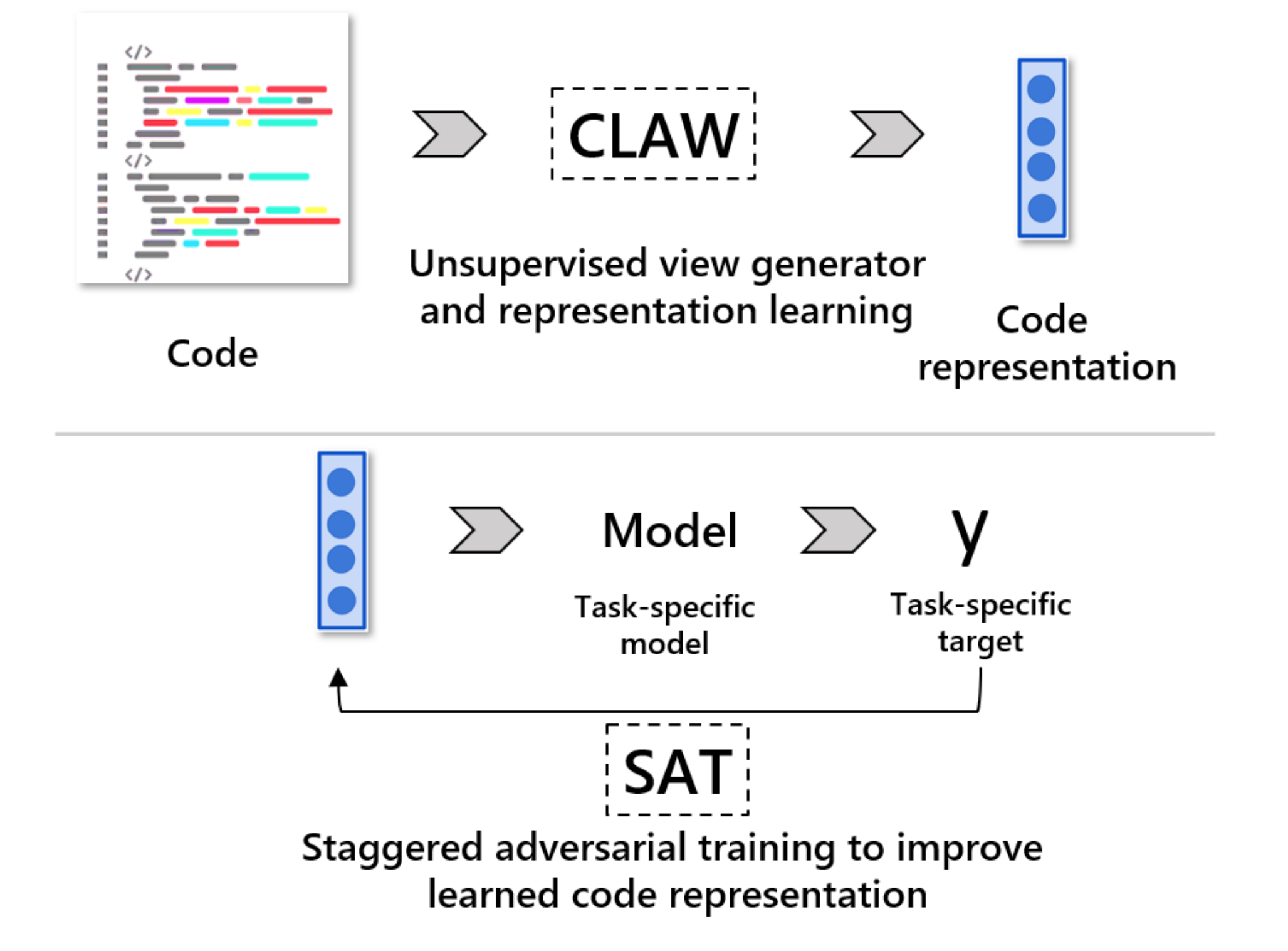}

  \caption{\textbf{Schematic overview.} We present \textbf{\claw} - a contrastive learning-based unsupervised method which learns \textit{adversarial views} of the input code to in turn learn accurate and robust representations of the code. We also present \textbf{SAT}, a refinement to the adversarial training algorithm proposed by Madry et al. \cite{madry2017towards} which helps retain the task-independent robustness and accuracy learned by \spc\claw while also learning task-specific accuracy and robustness. We show that \spc\ourmodel yields better accuracy and robustness when compared to state-of-the-art self-supervised learning models for code.}
  \label{fig: overview}
\end{figure}

We offer two methods that help us improve not only the transfer of robustness from pre-trained models to downstream tasks, but also co-improve fine-tuned accuracy and robustness. The schematic overview of our proposal is shown in Figure \ref{fig: overview}.
\textbf{First}, we propose a self-supervised pre-training method, \underline{c}ontrastive \underline{l}earning with \underline{a}dversarial vie\underline{w}s (\textbf{\claw}), which leverages adversarially-obfuscating codes as positive views of {\clsmall} so as to enforce the robustness of learned code representations. 
We formulate and achieve {\claw} through a bi-level optimization method.
We show that the representations learned from these pre-trained models yield better robustness transfer to downstream tasks.
\textbf{Second},  we propose \underline{s}taggered \underline{a}dversarial \underline{t}raining (\textbf{\sat}) to preserve the robustness learned during pre-training while also learning task-specific generalization and robustness during fine-tuning.  We show for the first time that the scheduler of adversarial code generation is adjustable and is a key to benefit both the generalization and robustness of code models.
\subsection{Contributions} \textit{On one hand}, we propose \spc\claw by extending the standard \spc\clsmall framework for code. 
As a baseline, we compare \spc\claw to the state-of-the-art standard \spc\clsmall framework for code models - \spc \textbf{\contracode} released by \cite{jain-etal-2021-contrastive}.
We find \spc\claw to `retain' more robustness when compared to \textbf{\contracode}.
Further, we provide a detailed analysis of understanding this improved performance from the perspective of model interpretability and characteristics of its loss landscape.
\textit{On the other hand}, we   integrate \spc\claw with \spc\sat to achieve   the {eventually fine-tuned} \textbf{\ourmodel} models. We evaluate three tasks--code summarization, code completion, and code clone detection, in two programming languages - Python and Java, and two different decoder models--LSTMs and transformers. We show that \spc\ourmodel outperforms {\contracode}  \cite{jain-etal-2021-contrastive} by roughly $6\%$ on the summarization task, $2\%$ on the completion task, and $1\%$ on the code clone task in accuracy, and by $9\%$, $3\%$, and $1\%$ on robustness respectively.
We study the effect of different attack strengths and attack transformations on this performance, and find it to be largely stable across different attack parameters. 

\section{Related work}
\label{sec:relatedwork}

Due to a large body of literature on SSL and adversarial robustness, we focus our discussion on those relevant to code models and CL (contrastive learning).

\subsection{SSL for code}
CL-based SSL methods offer a distinct advantage by being able to signal explicit examples where the representations of two codes are expected to be similar.
While the method itself is agnostic to the input representation, all the works in \spc\clsmall for code models work on code tokens directly.
Existing works \cite{jain-etal-2021-contrastive, chen2021varclr, bui2021, Wang2022SynCoBERTSM} show that  \spc\clsmall models for code improve the generalization accuracy of fine-tuned models for different tasks when compared to other pre-training methods like masked language models.
Each of these works uses semantics-preserving, {random code  
transformations} as positive views in its \spc\clsmall formulation. 
Such transformations help the pre-trained model learn the equivalence between representations of program elements which do not affect the executed output, such as the choice of variable names, the algorithmic `approach' used to solve a problem, \etc

State-of-the-art \spc\clsmall-based representations generally provide an improvement in the range of $1\%$-$5\%$   points of F1/BLEU/accuracy scores when compared to their fully supervised counterparts and other pre-training methods like transformers, which is significant in the context of the code tasks they evaluate, providing clear evidence for the utility of SSL methods. While these methods improve the generalizability of task-specific models, \textit{none of these works have studied the robustness of these models}, especially with a growing body of works showing the susceptibility of code models to adversarial attacks \cite{srikant2021generating,ramakrishnan2020,yefet2020adversarial}.
{
By contrast, the adversarial robustness of {\clsmall} models for image classification has increasingly been studied by the vision community \cite{fan2021does,jiang2020robust,kim2020adversarial,gowal2021selfsupervised}. 
These works have shown that {\clsmall} has the potential to offer dual advantages of robustness and generalization.
The fundamental differences in image and code processing, including how adversarial perturbations are defined in these two domains, motivate us to ask if and how the advantages offered by {\clsmall}  can be realized for code models.
}



\subsection{Adversarial robustness of code models: Attacks \& defenses} 
\cite{wang2019coset, quiring2019misleading, rabin2020evaluation,pierazzi2020intriguing} showed that obfuscation transformations made to code can serve as adversarial attacks on code models.
Following these works, recent papers \cite{yefet2020adversarial,ramakrishnan2020} proposed perturbing programs by replacing local variables and inserting print statements with replaceable string arguments. 
They found optimal replacements using a first-order optimization method, similar to HotFlip \cite{ebrahimi2017hotflip}.
\cite{srikant2021generating} framed the problem of attacking code models as a problem in combinatorial optimization, unifying the attempts made by prior works.
\cite{yefet2020adversarial, ramakrishnan2020, bielik2020adversarial} and \cite{srikant2021generating} also proposed strategies to train code models against adversarial attacks. While \cite{bielik2020adversarial} employed a novel formulation to decide if an input is adversarial, the other works employed the adversarial training strategy proposed by \cite{madry2018towards}.
Recently,\cite{yang2022natural} proposed a black-box attack method to generate adversarial attacks for code, which is different from the white-box setting used in \cite{wang2019coset, quiring2019misleading, rabin2020evaluation,pierazzi2020intriguing}. In this paper, we only focused on the adversarial robustness of white-box attacks. 
\subsec{Work most relevant to ours.} Our work comes closest to \contracode, the system proposed by \cite{jain-etal-2021-contrastive}. 
While they established the benefit of using \clsmall-based unsupervised representation learning for code, 
the work neglects  the interrelationship between pre-training and fine-tuning in the SSL paradigm, and the consequences of this relationship on both the robustness and generalization of the final model.
\section{Preliminaries}
\label{sec:motivation}
We begin by providing a brief background on code models, code obfuscation transformations, and  SSL-aided predictive modeling for code. 
We then motivate the problem of how to advance SSL for code models.
We study this through the lenses of accuracy and robustness of the learned models.

\subsection{Code and  obfuscation transformations} 
\begin{figure}[htb!]
    \centering
    \resizebox{0.45\textwidth}{!}{
    \includegraphics[width=\linewidth]{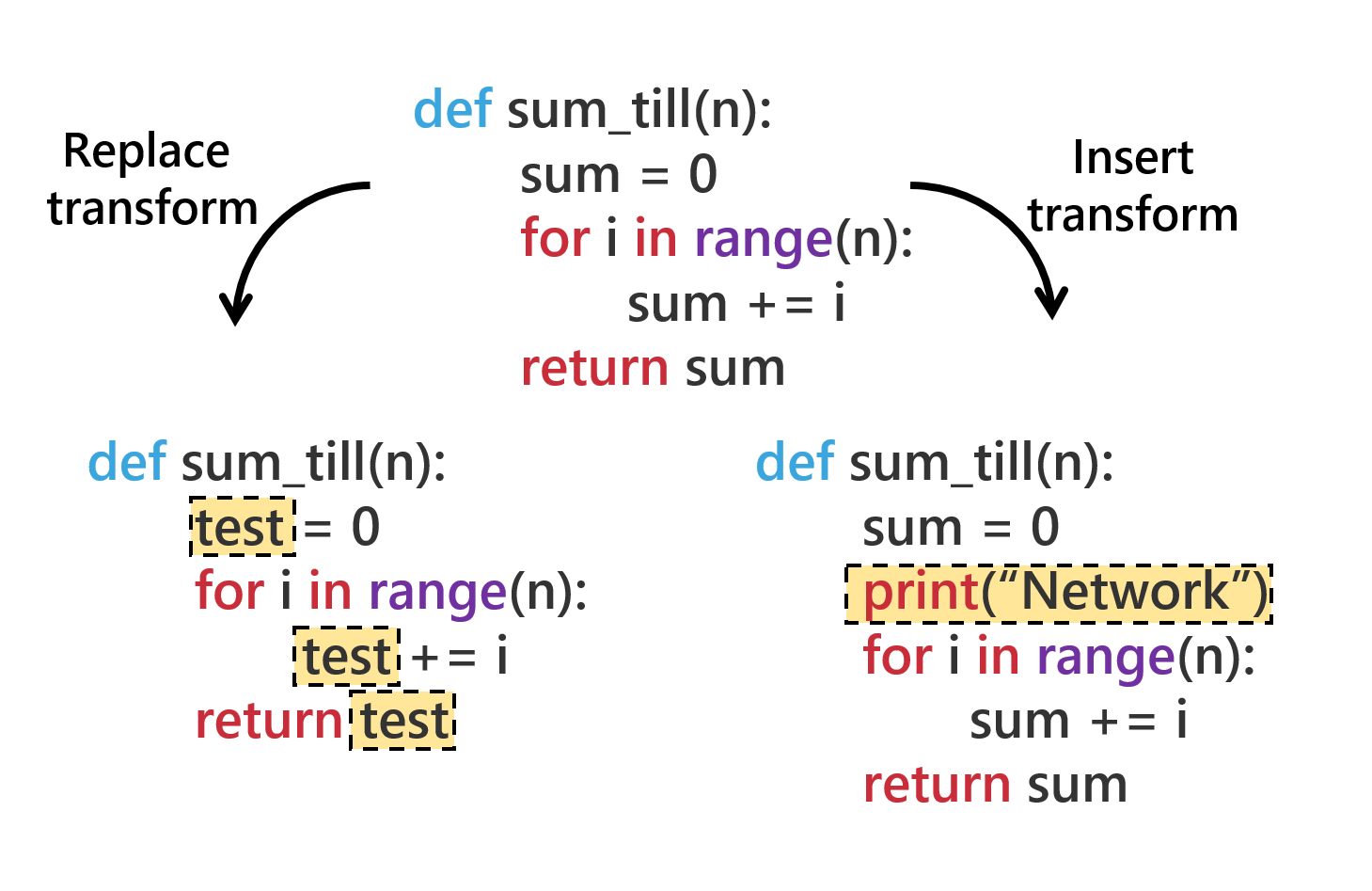}
    }
  \caption{{Two types of semantics-preserving transformations (obfuscations) can be made to a code to attack code models--{\spc\replace} - where existing code is modified at a \spc\site---location in the code, or {\spc\insertt} - where new lines of code are inserted at a \site. We select \spc\sites at random. The specific tokens used in these transformations (\texttt{test} and \texttt{"Network"} in the the example) can either be a random transformation $t_\text{rand}(\cdot)$--a randomly selected token from a pre-defined vocabulary, or can be an adversarial transformation $t_\text{adv}(\cdot)$, where the token is obtained from solving a first-order optimization designed to fool the model \cite{srikant2021generating,ramakrishnan2020,yefet2020adversarial}.}}
  \label{fig: adv_code_obf_examples}
\end{figure}

Let {\pnormal} denote a \textit{computer program} ({\ie}, code) which consists of a series of $n$ tokens $\{ \prog_i \}_{i=1}^n$ in the {source code domain}. For example, Figure \ref{fig: adv_code_obf_examples} shows an example code $\prog$.  
Given a  vocabulary of tokens (denoted by $\Omega$), each token can be regarded as a one-hot vector of length $|\Omega|$. 
Here we ignore white spaces and other delimiters when tokenizing.


Let $t(\cdot)$ denote an \textit{obfuscation transformation}, and $t(\prog)$ an obfuscated version of $\prog$.
$t(\prog)$ is semantically the same as $\prog$ while possibly being different syntactically.
Following the notations defined by \cite{srikant2021generating} and \cite{ramakrishnan2020},  we refer to locations or tokens in a code which can be {transformed} as \spc\sites.
We focus on \textit{replace} and \textit{\insertt} transformations, where either existing tokens in a source code are replaced by another token, or new lines of code are inserted in the existing code.
For example, in Figure\,\ref{fig: adv_code_obf_examples}, the \spc\replace transformation modifies the variable {\small\texttt{sum}} with {\small\texttt{test}}, while the \spc\insertt transformation introduces a new line of code {\small\texttt{print("Network")}}.

Obfuscation transformations have been shown to serve as adversarial examples for code models (see Section \ref{sec:relatedwork}).
During an adversarial attack, these transformations are made with the goal to get the resulting transformed code to successfully fool a model's predictions.
The transformations at any given \spc\site in a code, such as the tokens {\small\texttt{test}}, {\small\texttt{"Network"}} in Figure \ref{fig: adv_code_obf_examples}, can be obtained in two ways---by random transformations $t_\text{rand}(\cdot)$: they introduce a token sampled at random from $\Omega$, or through adversarial transformations $t_\text{adv}(\cdot)$: they solve a first-order optimization problem such that the transformed code {maximizes} the chances of the model making an {incorrect} prediction.

Our goal then turns to  improve  not only \textit{accuracy} (\ie prediction accuracy of properties of code $\prog$) but also \textit{robustness} (in terms of prediction accuracy of properties of $t(\prog)$, obfuscated transformations of $\prog$).

\subsection{Problem statement}
SSL typically includes two learning stages: 
\textit{self-supervised pre-training} and  \textit{supervised fine-tuning}, where the former acquires deep representations of input data, and the latter uses these learned features to build a supervised predictor specific to a downstream task, {\eg} code summarization \cite{alon2018general} as considered in our experiments. In the pre-training phase,
let $\boldsymbol \theta$ denote a feature-acquisition model (trained over   unlabeled data), and $\ell(\boldsymbol \theta)$ denote a pre-training loss, {\eg} the normalized temperature-scaled cross-entropy (NT-Xent) loss used in CL \cite{wu2018unsupervised,chen2020simple,he2020momentum}. 
In the supervised fine-tuning phase, let  $\boldsymbol \theta_{\mathrm{ft}}$  denote the prediction head appended to the representation network   $\boldsymbol \theta$, and  $\ell_{\mathrm{ft}}(\boldsymbol \theta_{\mathrm{ft}} \circ \boldsymbol \theta)$ denote
a task-specific fine-tuning loss seen as  a function of the entire model $\boldsymbol \theta_{\mathrm{ft}} \circ \boldsymbol \theta$, where $\circ$ denotes model composition.
Fine-tuning is performed over labeled data.   
The SSL pipeline can then be summarized as
%
\begin{align}
    \begin{array}{rl}
    \text{Pre-training:}     & \boldsymbol \theta_{\mathrm{pre}} = \displaystyle \arg\min_{\boldsymbol \theta} ~ \ell(\boldsymbol \theta),\\
    \text{(Full) Fine-tuning:}     & \displaystyle \minimize_{\boldsymbol \theta_{\mathrm{ft}}, \boldsymbol \theta} ~ \ell_{\mathrm{ft}}(\boldsymbol \theta_{\mathrm{ft}} \circ \boldsymbol \theta), ~ \\ &\text{with initialization $\boldsymbol \theta = \boldsymbol \theta_{\mathrm{pre}}$}.
    \end{array}
    \label{eq: pipeline_SSL}
\end{align} 
We remark that  if we fix $\boldsymbol \theta = \boldsymbol \theta_{\mathrm{pre}}$ in   \eqref{eq: pipeline_SSL}  during fine-tuning, then the resulting scheme is called \textit{partial fine-tuning} (\textbf{PF}), which only learns the prediction head $\boldsymbol \theta_{\mathrm{ft}}$.  
{Based on \eqref{eq: pipeline_SSL}} for code, this work tackles the following research questions:

\begin{tcolorbox}[
	standard jigsaw,
	opacityback=0,  
	]
\textit{(Q1) How to design a self-supervised pre-training scheme to acquire $\boldsymbol \theta_{\mathrm{pre}}$ that is robust to obfuscating codes?} 

\textit{(Q2) How to design a supervised fine-tuning scheme that can not only preserve the  generalization and robustness abilities gained from pre-training but also achieve new improvements via task-driven supervised learning?}
\end{tcolorbox}





\section{Method}
\label{sec:method}

{In this section, we will  study the   above (Q1)-(Q2) in-depth.}
To address  
 (Q1), we will develop a new pre-training method, termed {\claw}, which 
 integrates \underline{CL} with  \underline{a}dversarial \underline{v}iews of codes.
 The rationale is that
promoting the invariance of representations to possible adversarial candidates should then likely improve the robustness of models fine-tuned on these representations. 
To answer (Q2), we will   a novel fine-tuning method, termed \underline{s}taggered \underline{a}dversarial \underline{t}raining ({\sat}), which  can balance the supervised fine-tuning with unsupervised pre-training.
The rationale is that the supervised fine-tuning overrides pre-trained data representations and hardly retains the robustness and generalization benefits achieved during pre-training. 
We will show that the interplay between pre-training and fine-tuning should be carefully studied for robustness-generalization co-improvement in SSL for code. 


\begin{figure}[b!]
    \centering
    \resizebox{0.45\textwidth}{!}{
    \includegraphics[width=\linewidth]{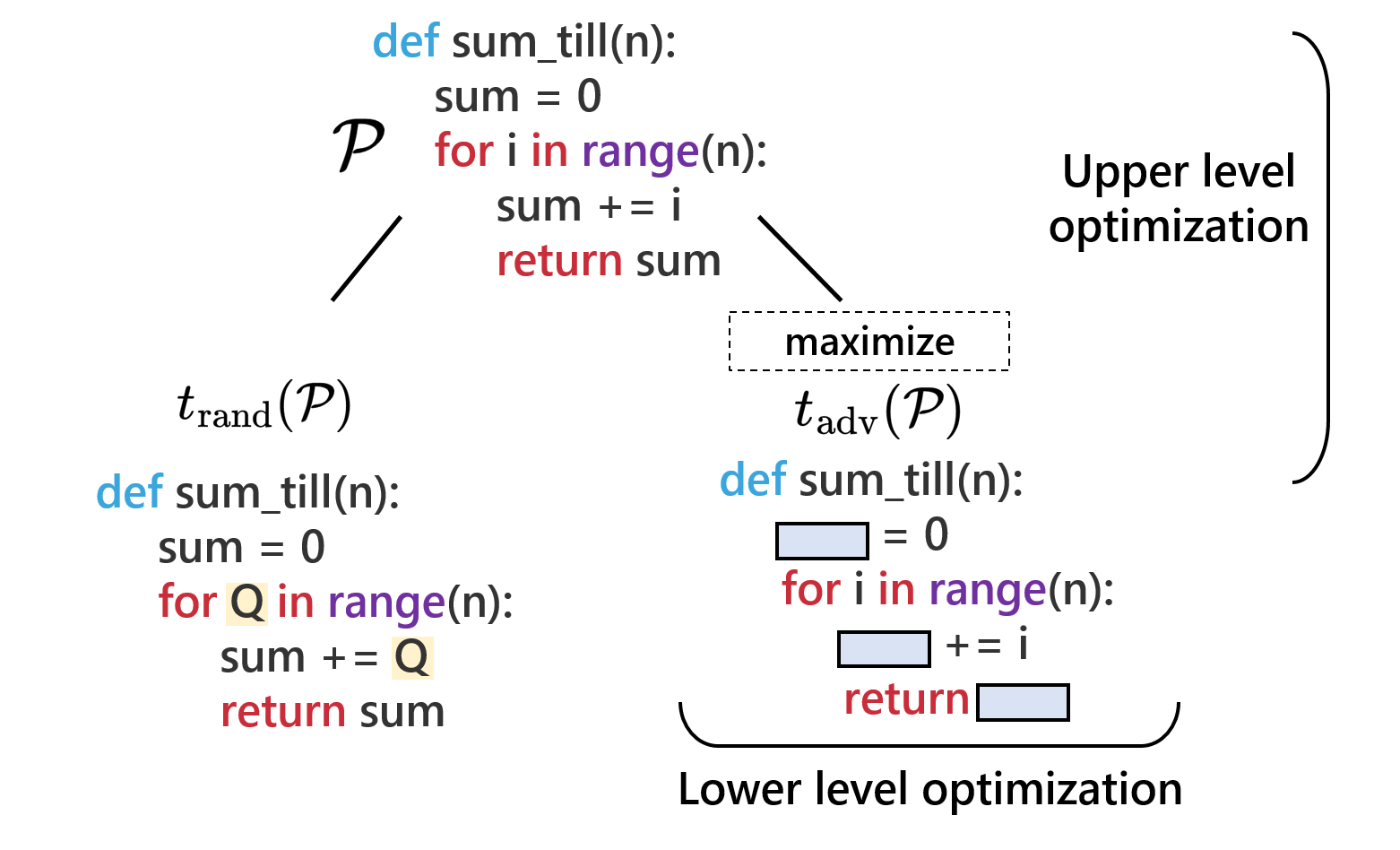}

    }
  \caption{{During pre-training, we propose \textbf{\claw} containing two optimization problems: (1) to {learn invariant code representations by minimizing the representation distances of a code ($\prog$) from all its views ($t_\text{rand}(\prog)$, $t_\text{adv}(\prog)$) via {\clsmall}}, and (2) to {generate an adversarial code $t_\text{adv}(\prog)$ (`hard' positive example) by  maximizing its representation distance from $\prog$}. In the example, this requires solving for a replacement token at the randomly selected \spc\site marked as \fcolorbox{black}{gblue}{\textbullet}.}}
  \label{fig: sat_example}
\end{figure}
\subsection{\claw: CL with adversarial codes}
CL \cite{chen2020simple,he2020momentum} proposes to first construct `positive' example pairs ({\ie}, original data paired with its transformations or  `views'), and then maximize agreement between them  while contrasting with the rest of the data (termed   `negatives').
In programming languages, code obfuscation transformations naturally serve as view generators of an input code. 
{While we reuse the same set of transformations (that are applicable to Python and Java programs) employed by the prior work \cite{jain-etal-2021-contrastive}, we generate transformed views of the code differently. \cite{jain-etal-2021-contrastive} use random views in their CL setup, which select and apply a transformation at random from the set of permissible transformations. 
We generate worst-case, optimization-based adversarial codes \cite{ramakrishnan2020, srikant2021generating} resulting in `adversarial views'.}

Infusing the original code, its random view, and its adversarial view  into CL, we obtain a \textit{three-view} positive tuple, denoted by ($\prog$, $t_{\text{rand}}(\prog)$, $t_{\text{adv}}(\prog)$). 
Since the generation of adversarial code (the right tokens for a given site) is in itself an additional optimization task, we leverage a bi-level optimization (BLO) framework \cite{liu2021investigating,zhang2022revisiting} and define optimization problems at two levels: the upper-level problem aims to solve the multi-view \spc\clsmall, while the lower-level problem aims to solve adversarial code generation (see an illustration   in Figure\,\ref{fig: sat_example}).
This results in the following formulation for our proposed approach {\claw}:
\begin{align}
  \begin{array}{l}
\displaystyle \minimize_{\boldsymbol \theta}  ~  \mathbb E_{\prog, t_\text{rand}}  [ \ell_{ \mathrm{NT-Xent}}(\boldsymbol \theta ; \prog, t_\text{rand} (\prog))  ]  + \\ \hspace*{16mm} \underbrace{\mathbb E_{\prog}  [ \ell_{\mathrm{NT-Xent}}(\boldsymbol \theta ; t_\text{rand}(\prog), t_\text{adv} (\prog) )  ]}_\text{Upper level: Multi-view CL}
 \vspace*{1mm}
 \\ \st ~ t_\text{adv} (\prog) = \underbrace{\displaystyle \arg\max_{\prog^\prime} \ell_{\mathrm{NT-Xent}}( \boldsymbol \theta; \prog, \prog^\prime ) }_\text{Lower level: Adversarial code generation},
    \end{array}
    \label{eq: BLO}
\end{align} 
where the three-view   objective function is constructed by    NT-Xent losses applied to two positive pairs $(\prog, t_\text{rand}(\prog))$ and $(t_\text{rand}(\prog), t_\text{adv} (\prog))$, respectively. {The first positive pair is   to gain the   generalizable code representation by promoting the representation invariance across the original view $\prog$ and   its (benign) randomly-obfuscating view  $t_\text{rand}$, as suggested in \cite{jain-etal-2021-contrastive}. The second positive pair is to enforce the adversary-resilient  code representation by promoting the representation invariance  across    the benign code   $t_\text{rand}$ and the adversarial code $t_\text{adv} (\prog)$.}
Given two codes $\prog_1$ and $ \prog_2$,   the specific form of $\ell_{\mathrm{ NT-Xent}}$ is given by as follows:
\begin{align}
  \begin{array}{l}
\ell_{\mathrm{ NT-Xent}}=-\frac{1}{2}\sum_{i=1}^2  \log   \frac{\exp\big(
\text{sim}(\mathbf z_{1}(\boldsymbol \theta), \mathbf z_{2}(\boldsymbol \theta))/t\big)}{ \sum_{k\in\mathcal N(i)},  \exp\big(\text{sim}(\mathbf z_{i}(\boldsymbol \theta), \mathbf z_k(\boldsymbol \theta))/t\big)}
    \end{array}
    \label{eq: Lnt}
\end{align} 
where  $\mathbf z_i$ denotes the 
feature representation of the input code $\prog_i$  achieved through the representation network $\boldsymbol \theta$, $\mathrm{sim}(\mathbf z_i, \mathbf z_j)$ denotes the cosine similarity between two feature representations $\mathbf z_i$ and $\mathbf z_j$, $t >0$   is a temperature parameter, and $\mathcal N(i)$ is the set of batch data except the data sample $i$ \cite{chen2020simple}. 

To solve the BLO problem \eqref{eq: BLO}, we apply an alternating optimization method \cite{bezdek2003convergence,liu2021investigating}. Specifically, by fixing the representation network $\boldsymbol \theta$,  the lower-level adversarial code generation  is accomplished using first-order gradient descent following \cite{srikant2021generating, ramakrishnan2020}. 
Given the generated adversarial code $t_{\mathrm{adv}}(\mathcal P)$, we then in turn solve the upper-level CL problem. The above procedure is alternatively executed for every data batch.


\subsec{Adversarial view is beneficial to representation learning.}
To highlight the effectiveness of incorporating adversarial codes in CL at the pre-training phase,
the rows `{\contracode-PF}' and `{\claw}-PF' of
 Table\,\ref{tab:pf-robustness} demonstrate a warm-up experiment by  comparing the  performance of the proposed pre-training method {\claw} with that of the baseline approach \spc\contracode \cite{jain-etal-2021-contrastive}.
To precisely characterize the effect of the learned representations on code model generalization and robustness, we partially fine-tune (PF) a \spc\seqtseq model on the downstream task of summarizing code (details in Section \ref{sec:experiments}) in Python (\textbf{\summarypy}) and Java (\textbf{\summaryjava}) by fixing the set of weights learned during pre-training.
\textbf{{\genf}} and \textbf{{\robf}} are the generalization F1-scores and the robust F1-scores, {\ie}, F1 scores of the model when attacked with adversarial codes. 
Partially fine-tuning these models allows us to study the sole contribution of the pre-training method in the learned robustness and generalization of the model.
As we can see, the partially fine-tuned {\claw} model (termed {\claw-\pf}) outperforms the baseline {\contracode}-{\pf}, evidenced by the substantial robustness improvement 
(4.38\% increase in {\robf} in \summarypy) as well as lossless or better generalization performance (1.58\% increase in \spc\genf scores on \summaryjava).  
It is worth noting that 
adversarial codes serve as `hard' positive examples in the representation space (given by maximizing the representation distance between $\prog$ and its perturbed variant $\prog^\prime$ in the lower optimization level of {\claw},  \eqref{eq: BLO}). The benefit of hard positive examples in improving generalization has also been seen in vision \cite{chuang2020debiased,wang2020unsupervised,fan2021when}.

\begin{table}[htb]
\renewcommand{\arraystretch}{1.4}
\begin{threeparttable}
\resizebox{0.45\textwidth}{!}{
\begin{tabular}{c|c|c|c|c}
\hline
\multirow{3}{*}{\begin{tabular}[c]{@{}c@{}}Model\end{tabular}} & \multicolumn{4}{c}{\textbf{Partial fine-tuning }} \\ 
\cline{2-5} & \multicolumn{2}{c|}{\textbf{\summarypy}} & \multicolumn{2}{c}{\textbf{\summaryjava}}\\
\cline{2-5}
~&\multicolumn{1}{c|}{\textbf\genf} & \multicolumn{1}{c|}{\textbf\robf} & \multicolumn{1}{c|}{\textbf\genf} & \multicolumn{1}{c}{\textbf\robf}\\
\hline
        {\small \notate{\contracode}{\pf}{\pf}}
         & 25.46  &  15.47 & 20.92 & 16.63 \\
        \notate{\claw}{\pf}{\pf}
        & 25.45  &  19.05 & 22.50 & 17.14  \\
        \hline
        ~&\multicolumn{4}{c}{\textbf{Full fine-tuning }} \\
        \hline
          \notate{\contracode}{\ff}{\stdtr} & \multicolumn{1}{c|}{36.28}  & \multicolumn{1}{c|}{28.97}  & \multicolumn{1}{c|}{41.37}   & \multicolumn{1}{c}{33.01} \\
     \notate{\claw}{\ff}{\stdtr} & 36.57   & 29.97  & 41.23 & \multicolumn{1}{c}{32.53 }    \\
       \notate{\contracode}{\ff}{\advtr} & 32.80   & 32.39     & 38.67   & \multicolumn{1}{c}{35.91}    \\
      \notate{\claw}{\ff}{\advtr} & 32.97   & 32.65 &  38.86   & \multicolumn{1}{c}{36.10}    \\
      \hline
\end{tabular}

}

\end{threeparttable}
\caption{{{Partially fine-tuned (PF) models show that \spc\claw improves robustness. Standard training (\stdtr) yields better generalization than adversarial training (\advtr) while the latter provides better robustness.  
}}}
\label{tab:pf-robustness}
\end{table}





\subsection{{\sat}: Staggered adversarial training for fine-tuning}

As shown in the previous section, an appropriate pre-training method can improve the quality of learned deep representations, which help improve the robustness and accuracy of a code model. 
However, the state of the model present at the end of the pre-training phase may no longer hold after supervised fine-tuning. 
That is because supervised learning (trained on labeled data vs. unlabeled data in representation learning) may significantly alter the characteristics of the learned representations. 
Thus, a desirable fine-tuning scheme should be able to yield accuracy and robustness improvements \textit{complementary} to the representation benefits provided by pre-training. 
Towards this goal, we posit that fine-tuning should \textit{not} be designed in a way which merely optimizes a single performance metric--either accuracy or robustness. 

To justify this hypothesis, we consider two extreme cases during fine-tuning: \textit{(i)} standard training (ST)-based {\ff}, and \textit{(ii)} adversarial training (AT)-based {\ff} \cite{madry2018towards}.
\spc\stdtr is essentially the same setup as fully supervised training with the only difference being in the set of initial parameters of the model.
This setup optimizes improving a model's generalization ability.
On the other hand, \spc\advtr optimizes improving the model's adversarial robustness.

The \textit{last four rows of Table\,\ref{tab:pf-robustness}} present the performance of these two extreme fine-tuning cases applied to the pre-trained models provided by {\contracode}  \cite{jain-etal-2021-contrastive}  and {\claw}, respectively. 
As we can see, when either {\stdtr} or {\advtr} is used, different pre-training methods ({\contracode} and {\claw}) lead to nearly the same generalization and robustness performance.
This shows that fine-tuning, when aggressively optimizing one particular performance metric, could override the benefits achieved during pre-training.
To this end, we propose staggered AT ({\sat}), a hybrid of {\stdtr} and {\advtr} by adjusting the time instances (in terms of epoch numbers) at which adversarial codes are generated (see Algorithm \ref{alg:sat}).
\loadFig{alg:sat}
{\sat} involves two key steps---training a model $\mathcal{M} \Def \{ \boldsymbol \theta, \boldsymbol \theta_{\mathrm{ft}} \}$ on a batch $\mathcal{B}$ of data (step 1), and attacking the learned model at a staggered frequency $\tau$ (steps 2-3).
\textit{Different from {\advtr}}, adversarial code is not generated in every data batch. 
Instead, in {\sat}, we propose reducing the frequency of adversarial learning. 
Accordingly, adversarial code generation occurs at the frequency of each epoch or at every few epochs.
This ensures the model parameters retain as much of the attributes from pre-training while also learning task-specific generalization and robustness.
In \sat, the model is finetuned  using ($\mathcal{B}_i$ + $\mathcal{B}^\prime_i$) where $\mathcal{B}_i^\prime$ refers to the generated adversarial code corresponding to $\mathcal{B}_i$.
Eventually,  by combining the proposed pre-training scheme {\claw} with the fine-tuning scheme {\sat}, we term the resulting SSL framework for code as {\ourmodel}.
\section{Experiment Setup}
\label{sec:experiments}
We describe the following aspects of our experiment setup - the task, dataset, and the details of the model.

\loadFig{tab:generalization-robustness}
\subsec{Task, dataset, error metrics.} We evaluate our algorithm on \textbf{four} tasks: \textbf{(1)} code summarization \cite{alon2018general, aloncode2vec, allamanis2018learning, wang2020learning, david2020neural, jain-etal-2021-contrastive} in Python, \textbf{(2)} code summarization in Java (generates English description for given code snippet), \textbf{(3)} code completion in Python \cite{lu2021codexglue} (generates the next six tokens for a given code snippet), and \textbf{(4)} code clone detection \cite{wang2020detecting} in Java (classifies whether a pair of code snippets are clones of each other).
For models evaluated in Python, we pre-train on the \spc\pylarge dataset \cite{csn}, containing $\sim$~$500$K methods, and fine-tune on the \spc\pysmall dataset \cite{sripy}, containing $\sim$~$200$K methods.
For Java, we pre-train on the \spc\javalarge dataset \cite{csn} containing $\sim$~$600$K and fine-tune on the \spc\javasmall \cite{alon2018general} dataset containing $\sim$~$500$K methods.
We use the F1-score $\in[0,100]$ to measure the performance of all our models, consistent with all the related works, including \cite{jain-etal-2021-contrastive}.
A higher value indicates that the model generalizes better to the task.
While these F1-scores are correlated to BLEU scores, they directly account for token-wise mis-predictions. The F1 scores are computed following \cite{allamanis2016convolutional,alon2019code2vec}
Specifically, for each of our models, we reuse the two F1 scores: \textbf{\genf} - the model's generalization performance on a task, and \textbf{\robf} - the model's performance on the task when semantics-preserving, adversarially-transformed obfuscated codes are input to it; see details below. 

\subsec{Adversarial attacks, attack strength, code transformations.}
When attacking code models, we use the formulation by  \cite{srikant2021generating} to define the strength of an attack. 
Specifically, selecting a larger number of \spc\sites in a code---locations or tokens in a code which can be \textit{transformed} to produce an adversarial outcome---corresponds to a stronger adversarial attack, since this allows multiple changes to be made to the code.
Also, we follow \cite{srikant2021generating,ramakrishnan2020} to specify the set of code transformations: 
\spc\replace (renaming local variables, renaming function parameters, renaming object fields, replacing boolean literals) and \spc\insertt (inserting print statements, adding dead code).

In our setup, we can leverage the attack strength at three stages--during pre-training (using adversarially attacked code as views), when fine-tuning with adversarial training {\advtr} \cite{madry2018towards} or {\sat}, and when evaluating robustness on an unseen test set.
Unless specified otherwise, we  pre-train on one {\site}, attack one \spc\site in each iteration of \sat, and attack one {\site} during evaluation.
In Table \ref{tab:sensitivity} 
,we analyze the effect of varying the number of sites at each of these stages.
We apply the first-order  optimization method proposed in \cite{ramakrishnan2020} 
to generate adversarial codes.
To adversarially train these models during fine-tuning, we employ either {\advtr}  \cite{madry2018towards} or our proposed {\sat} for code.

\subsec{Baselines.} We compare \spc\ourmodel to \textbf{three} baselines.
\textbf{(1)}
A {supervised model} (model \textbf{\modelnum{1}} in Table \ref{tab:generalization-robustness}) - Pre-training has no effect on a fully supervised model.
\textbf{(2)} {Adversarially trained supervised model} (\textbf{\modelnum{2}}) {on top of $\modelnum{1}$} - we use the \spc\advtr setup first proposed by \cite{ramakrishnan2020}, which in turn employs the setup from \cite{madry2018towards}.
Due to the characteristics of \advtr, we expect to see an improvement in its \spc\robf as compared to \modelnum{1} but a decrease in \genf.
\textbf{(3)}
The \spc\contracode model (\textbf{\modelnum{3}}) from \cite{jain-etal-2021-contrastive}.
\spc\contracode reflects the state-of-the-art in pre-training methods as it outperforms other pre-training models like BERT-based models and GPT3-Codex. 
Hence, we do not compare ourselves again to other pre-training models.

\subsec{Models.} 
For the summarization and completion tasks, we experiment with two seq2seq architectures--LSTMs and transformers.
For the detection task, we use a fully connected linear layer as a decoder.
The decoders are trained to predict the task (generating English sentence summaries, generating code completions, flagging code clones respectively) in both the fine-tuned and standard training settings.
When fine-tuning, we use the learned encoders from the pre-trained models.
For the summarization and completion tasks, we report all our results on the LSTM decoder (Table \ref{tab:generalization-robustness}), and compare the performance of transformers in our ablation study. The LSTM encoder has 2 layers across all experiments. In the code summarization tasks, there exists another two-layer decoder added to the encoder to generate the summary of the programming   language. In the code completion task, we also add a two-layer decoder to generate the code snippets. In the code clone detection task, we  add one linear layer to map the data representations to the data labels. As for the transformer architecture, The transformer encoder has 6 layers and the transformer decoder has another 6 layers to generate the programming language summary. 


\DeclareFixedFont{\ttb}{T1}{txtt}{bx}{n}{10} 
\DeclareFixedFont{\ttm}{T1}{txtt}{m}{n}{10}  

\definecolor{RoyalBlue}{cmyk}{1, 0.50, 0, 0}
\lstset{language=Java,
    keywordstyle=\color{RoyalBlue},
    basicstyle=\scriptsize\ttm,
    commentstyle=\ttfamily\itshape\color{gray},
    stringstyle=\ttfamily,
    showstringspaces=false,
    breaklines=true,
    frameround=ffff,
    autogobble=true,
    numbers=none
}
\definecolor{deepblue}{rgb}{0,0,0.5}
\definecolor{deepred}{rgb}{0.6,0,0}
\definecolor{deepgreen}{rgb}{0,0.5,0}
\newcommand\pythonstyle{\lstset{
		language=Python,
		basicstyle=\ttm,
		breaklines=true,
		postbreak=\mbox{\textcolor{red}{$\hookrightarrow$}\space},
		otherkeywords={self},             
		keywordstyle=\ttb\color{deepblue},
		emph={MyClass,__init__},          
		emphstyle=\ttb\color{deepred},    
		stringstyle=\color{deepgreen},
		showstringspaces=false            %
	}}
\newcommand\pythonexternal[2][]{{
		\pythonstyle
		\lstinputlisting[#1]{#2}
}}

\subsec{Hyperparameter setup.}
We optimize model parameters   using Adam with linear learning rate warm-up. 
For the bidirectional LSTM encoders, the maximum learning rate for {\contracode} and {\claw} is $10^{-4}$, and then is decayed accordingly. For the transformer-type encoder, the maximum learning rate is $10^{-4}$ for  {\contracode} and $10^{-5}$ for {\claw}. For different downstream tasks, we optimize the parameters using Adam with step-wise learning rate decay. The maximum learning rates of \stdtr, \advtr, {\sat} are $10^{-3}$ on code summarization and code completion tasks, and $10^{-4}$ for code clone detection tasks. For downstream tasks using transformer, the learning rates of \stdtr, \advtr, {\sat} are $10^{-4}$. All of the downstream tasks are finetuned for 10 epochs with practical convergence, and we utilize a validation dataset to pick the best-performed model. All of the experiments are conducted on 4 Tesla V100 GPU with 16 GB memory.

\section{Experiment Results}

We summarize the overall performance of \spc\ourmodel and follow that up with multiple additional analyses and ablations to better understand our model's performance.

\subsection{Overall performance}
We evaluate the different pre-training and fine-tuning strategies we consider in Section\,\ref{sec:method}.
The accuracy and robustness F1-scores of the different models are shown in Table \ref{tab:generalization-robustness}.
Models \modelnum{1}-\modelnum{3} are the baselines described in the previous section.
Model \modelnum{4} pertains to using a {\claw} encoder with standard training (ST) for the downstream task.
Model \modelnum{5} pertains to using a {\claw} encoder with standard adversarial training (AT) for the downstream task.
And finally, \modelnum{6} pertains to using {\sat} for the downstream task with a {\claw} encoder. We observe the following:

\textbf{First}, from the perspective of accuracy, Table \ref{tab:generalization-robustness} shows that our proposal \spc\ourmodel (model \modelnum{6}) outperforms all the baselines on \genf.  
Particularly, {\ourmodel} achieves nearly 8\% accuracy improvement over {\contracode} (\modelnum{3}) on \summarypy.
\textbf{Second}, we see that \spc\ourmodel can substantially help improve the robustness (measured by {\robf}) of a downstream model. 
Compared to \modelnum{3}, we see an improvement of $14.7\%$ for \summarypy, and $5.8\%$ for {\summaryjava}. 
Similarly, the gain in robustness is much more substantial in {\completepy} than in accuracy.
{\clonejava} is the simplest of the four tasks, and we see comparable accuracies across all the models we evaluate.
\textbf{Additionally}, we adversarially train baselines as well (models \modelnum{2} and \modelnum{5} respectively) and compare their robustness scores to ours  (\modelnum{6}).
We make two observations--\begin{inparaenum}[(a)]
\item As is expected with \advtr, we notice a drop in these models' accuracy--the \spc\genf scores of \modelnum{5} is lower than that of its standard-training counterpart \modelnum{4} (the same trend is observed between \modelnum{2} and \modelnum{1} as well).
\item While {\advtr}-based fine-tuning provides an expected improvement in robustness over the other baselines that use {\stdtr}-based fine-tuning, the {\robf} they achieve is still much lower than our model. 
This is because the robustness gain at the pre-training phase was overridden by {\advtr} at the fine-tuning phase.
\end{inparaenum}
\begin{tcolorbox}[
	standard jigsaw,
	opacityback=0,  
	]
In summary, \textbf{Table \ref{tab:generalization-robustness}} shows that the proposed \spc\ourmodel allows us to learn task-specific accuracy and robustness while preserving these attributes learned during pre-training.
\end{tcolorbox}
In what follows, we analyze the performance of \spc\claw and \spc\sat separately from different perspectives.



\subsection{Why is {\claw} effective? A model landscape perspective}
\loadFig{lossclaw}
\begin{figure*}[t]
\centerline{
\renewcommand{\arraystretch}{1.2}
\resizebox{1.1\textwidth}{!}{
\begin{tabular}{l  l} 
\pythonexternal{codes/1.py} &  
\pythonexternal{codes/3.py} \\
\multicolumn{1}{c}{\small{(a) Sample}} & \multicolumn{1}{c}{\small{(b) Adversarially perturbed version of sample (a1)}} \\

\pythonexternal{codes/4.py} & 
\pythonexternal{codes/4.py} \\
\multicolumn{1}{c}{\small{(c) EBE similar to (a) - \claw}} & \multicolumn{1}{c}{\small{(d) EBE similar to (a) - \contracode}} \\
\pythonexternal{codes/4.py} & 
\pythonexternal{codes/2.py} \\
\multicolumn{1}{c}{\small{(e) EBE similar to (b) - \claw}} & \multicolumn{1}{c}{\small{(f) EBE similar to (b) - \contracode}}  \\


\end{tabular}
}
}
\caption{{Explanation-by-example to demonstrate the robustness benefits of {\claw}. \textbf{(a)} Sample program from the test set \textbf{(b)} Adversarially perturbed variant of the sample program.
\textbf{(c-d)} Examples closest to the sample program (a) when using  {\claw} and \contracode. 
\textbf{(e-f)}   Examples closest to the perturbed variant (b) when using {\claw} and \contracode. 
}} 
  \label{fig: ebe}
\end{figure*}
Liu \etal \cite{liu2020towards} show that the generalization benefit of an approach in the`pre-training + fine-tuning' paradigm  can be deduced by the flatness of the loss landscape of the pre-trained model.
This would then let fine-tuning to force the optimization for fine-tuning to stay in a certain neighborhood of the pre-trained model of high quality.

To plot the loss landscapes, we follow the procedure from Li \etal \cite{li2018visualizing} by plotting 
\begin{align}
f(\alpha,\beta) = \ell(\theta^\star+\alpha\delta+\beta\eta)
\label{eq:losslandscape}
\end{align}
where $\delta$ and $\eta$ are two random direction vectors in the parameters' subspace, and $\theta^\star$ is the parameters of a model.
We average the supervised loss from the partially fine-tuned models from $640$ randomly selected samples in our test set (out of $10000$, $6.4\%$). 

\begin{tcolorbox}[
	standard jigsaw,
	opacityback=0,  
	]
Results of \textbf{Figure \ref{fig:lossclaw}} confirm the \textbf{flatness} of the loss landscape in \spc\claw when compared to \contracode, implying a better transfer of generalizability by \claw.
\end{tcolorbox}


Next, we show 
another way to justify the flatness merit of {\claw}'s loss landscape. The key idea is to
  track the deviations of the weights of the pre-trained encoders in the fine-tuned setting, as  inspired by \cite{liu2020towards}. Specifically,  
let $\boldsymbol \theta_{\mathrm{pre}}$ and $\boldsymbol \theta_{\mathrm{pre}}^\prime$ denote the weights of the representation model $\boldsymbol \theta$ pre-trained by {\claw} and the fine-tuned weights obtained by using different fine-tuning methods, respectively. 
It was shown in \cite{liu2020towards}   that the generalization benefit of an approach in the`pre-training + fine-tuning' paradigm  can be deduced by the deviation between the fine-tuned weights $\boldsymbol \theta_{\mathrm{pre}}^\prime$ and the pre-trained weights $\boldsymbol \theta_{\mathrm{pre}}$. This would then let fine-tuning to force the optimization of $\boldsymbol \theta_{\mathrm{pre}}^\prime$ to stay in a certain neighborhood of $\boldsymbol \theta_{\mathrm{pre}}$. We have already shown that {\claw}  will lead to a flatter loss landscape compared to \contracode previously. 
Here we computed the Frobenius norm $||\boldsymbol \theta_{\mathrm{pre}}^\prime-\boldsymbol \theta_{\mathrm{pre}}||_{\mathrm{F}}$ as a proxy to justify the  generalization benefit following \cite{liu2020towards}.

\loadFig{tab:weights-epoch-model}

Table\,\ref{tab:weights-epoch-model} summarizes the aforementioned weight characteristics for the code summarization task.
As we can see, the weight deviation corresponding to {\claw} is less than that associated with {\contracode} given a finetuning method.  

\begin{tcolorbox}[
	standard jigsaw,
	opacityback=0,  
	]
The results from \textbf{Table\,\ref{tab:weights-epoch-model}} suggest that an encoder pretrained using {\claw} transfers better than that using {\contracode}. 
\end{tcolorbox}


\subsection{Interpretability of learned code representations}

We evaluate the   robustness benefit of \spc\claw through the lens of (input-level) model explanation.
Following the observations from Jeyakumar \etal \cite{explainbyexample} on probing models locally, we investigate \spc\claw and \spc\contracode using a training data-based model explanation method: explanation-by-example (EBE) \cite{kim2016examples}. 
The core idea is to leverage train-time data to explain test-time data by matching their respective representations. 
If the pre-trained models are robust, adversarially perturbing the samples should not alter their representations and thus should be mapped to the same set of closest training examples that were found without   perturbations.

\begin{tcolorbox}[
	standard jigsaw,
	opacityback=0,  
	]
Based on EBE, we sample {$100$} code snippets $\{\mathcal{P}^{\mathrm{test}}_i\}_{i=1}^{100}$ at random from our test dataset, and find the closest samples $\{\mathcal{P}^{\mathrm{train}}_{\claw}\}$ and $\{\mathcal{P}^{\mathrm{train}}_{\contracode}\}$ in the training dataset using the EBE method, based on representations produced by $\boldsymbol \theta_{\claw}$ and $\boldsymbol\theta_{\contracode}$ respectively.
We find that $68\%$ of the representations from \spc\claw match their original codes in $\{\mathcal{P}^{\mathrm{train}}_{\claw}\}$  while $57\%$ of \spc\contracode representations match their original codes in $\{\mathcal{P}^{\mathrm{train}}_{\contracode}\}$. \textbf{The above results} suggest that the learned representations by \spc\claw are more adversarially robust than \spc\contracode. 
\end{tcolorbox}


In what follows, we peer into the EBE method's results with an example below.
\subsec{$\bullet$ A sample program from the test-set:}
\vspace*{3mm}
\pythonexternal{codes/9.py}
\vspace*{3mm}
\subsec{$\bullet$ Adversarially perturbed variant of the sample program:}
The adversarial attack algorithm replaces the method argument \texttt{helper} with \texttt{edges}. 
\vspace*{3mm}
\pythonexternal{codes/10.py}
\vspace*{3mm}
\subsec{$\bullet$ EBE sample in the train-set closest to the sample program when using {\contracode} or {\claw}:}
This is the example whose {\contracode} or {\claw} representation (encoder trained by {\contracode} or {\claw}) is closest to the representation of the sample program. We can find that they have the same functionality. 
\vspace*{3mm}
\pythonexternal{codes/11.py}
\vspace*{3mm}
\subsec{$\bullet$ EBE sample in the train-set closest to the perturbed variants of sample program from {\contracode}:} We can observe that the closest program of the perturbed program in the training dataset is different from that of the original program. The new closest program has different functionality from the previous test sample program.
\vspace*{3mm}
\pythonexternal{codes/12.py}
\vspace*{3mm}
\subsec{$\bullet$ EBE sample in the training-set closest to the perturbed sample program from {\claw}:}
This example pertains to the representation that is closest to the representation of the sample program computed by an encoder trained by {\claw}. 
We see that despite comparing it to a perturbed sample's representation, the example found by EBE corresponds to the unpertrubed sample program, suggesting the robustness of {\claw} over \contracode.
\vspace*{3mm}
\pythonexternal{codes/11.py}
\vspace*{3mm}

We also provide more examples in Figure \ref{fig: ebe}.
Figure \ref{fig: ebe}.a shows a sample Python program and Figure \ref{fig: ebe}.b shows its respective adversarially perturbed variant.
The closest training programs in the training set mapped to the representations before perturbations are shown in Figures \ref{fig: ebe}.c and  \ref{fig: ebe}.d; and those mapped to the representations after perturbations are shown in Figures \ref{fig: ebe}.e and  \ref{fig: ebe}.f.

\begin{tcolorbox}[
    standard jigsaw,
 	opacityback=0,  
 	]
 	As shown in \textbf{Figure \ref{fig: ebe}},
we find that EBE consistently finds the same training examples for \spc\claw (Figure \ref{fig: ebe}.c and Figure \ref{fig: ebe}.e) irrespective of the adversarial perturbations made to the sample program, confirming its enhanced robustness. 
\end{tcolorbox}

\subsection{{\sat} enables generalization-robustness sweet spot}
\begin{figure}[h]
    \centering
    \hspace*{-0mm}\includegraphics[width=.45\textwidth,height=!]{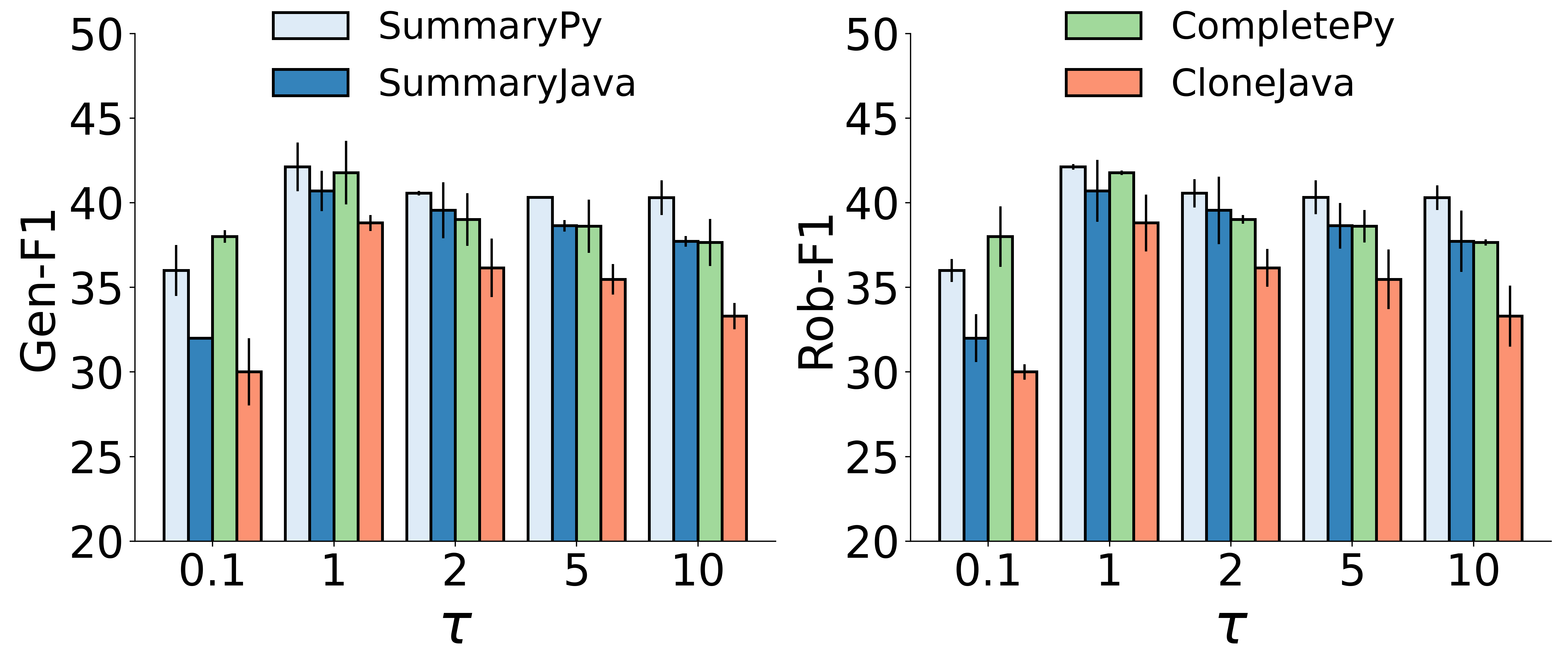}
    \caption{\small{Effect of different update schedules ($\tau$, see Algorithm \ref{alg:sat}) on \spc\genf and \robf.}
    }
\label{fig:robustness-accuracy-epochs}
\end{figure}

Figure \ref{fig:robustness-accuracy-epochs}.(A) shows the results from our experiments on the code summarization task where we vary $\tau$, the frequency of attacking the code model during
{\sat} (see Algorithm \ref{alg:sat}).

We generate adversarial code tokens every $\tau^\text{th}$ epoch, where we vary $\tau$ from less than $1$ (corresponds to an update occurring at every batch within an epoch; this is the \spc\advtr algorithm from \cite{madry2018towards}) to $10$.
The X-axis shows this frequency.
We plot both \spc\genf (left) and \spc\robf (right) of the adversarially trained model when varying $n$.
Across the four tasks we evaluate in this work, We find that a sweet spot exists in a less frequent epoch-wise schedule, associated with {\ourmodel} (\modelnum{6}, which corresponds to $\tau=1$), which improves both \spc\genf and {\robf} over \modelnum{5} (which corresponds to $\tau=0.1$).

\begin{tcolorbox}[
	standard jigsaw,
	opacityback=0,  
	]
Results of \textbf{Figure \ref{fig:robustness-accuracy-epochs}}   validate our hypothesis of being able to retain the robustness learned during pre-training while updating the model just enough during fine-tuning to `learn' new robustness while also learning the downstream task.
\end{tcolorbox}


\subsection{{\ourmodel} on a different architecture}
\loadFig{tab:transformer}
We  further evaluate {\sat} on a different model architecture.
We consider the transformer architecture (6-layer encoder and 6-layer decoder following \cite{jain-etal-2021-contrastive}), and observe similar results (see Table \ref{tab:transformer}): \spc\ourmodel offers the best accuracy and robustness.

\begin{tcolorbox}[
	standard jigsaw,
	opacityback=0,  
	]
\textbf{Table \ref{tab:transformer}} shows that {\ourmodel} performs well across multiple encoder architectures.
\end{tcolorbox}
\subsection{Extended study to integrate {\sat} with {\contracode}}
\loadFig{tab:generalization-robustness2}
To further verify the effectiveness of \sat, we employ it in \contracode---we modify their implementation to introduce a staggered adversarial training schedule. Table \ref{tab:generalization-robustness2} tabulates its performance.
We find that {\spc\sat} also benefits \spc\contracode, but the gain is smaller than \spc\ourmodel (\modelnum{6}, Table\,\ref{tab:generalization-robustness}). 
This justifies the complementary benefits of  {\claw} and {\sat}.
\begin{tcolorbox}[
	standard jigsaw,
	opacityback=0,  
	]
\textbf{Table \ref{tab:generalization-robustness2}} shows the complementary benefits of {\claw} and {\sat} on the state-of-the-art SSL method {\contracode} as well, demonstrating the effectiveness of the two model-independent techniques we introduce in this work.
\end{tcolorbox}
\subsection{Sensitivity of {\sat} to  code transformation and attack strength types.}
\loadFig{tab:sensitivity}
We evaluate the sensitivity of our best-performing model on differing attack conditions. 
We consider two factors: (1) Transformation type: we study the effect of the two transformation types--\spc\replace and \insertt, and their combination. (2) Attack strength: we vary the number of \sites---locations in the codes that can be adversarially transformed.

We summarize our results in Table \ref{tab:sensitivity}.
The values \texttt{a}, \texttt{b} in each cell correspond to \spc\genf and \spc\robf of \spc\ourmodel respectively.

\begin{tcolorbox}[
	standard jigsaw,
	opacityback=0,  
	]

The results from \textbf{Table \ref{tab:sensitivity}} suggest that when using transformations in pre-training or in fine-tuning, it is advisable to use a combination of both \spc\replace and \spc\insertt transformations.
When evaluating \ourmodel's robustness against stronger adversarial attacks, we find  \spc\robf consistently outperforms {\contracode} in all the configurations we evaluate.
\end{tcolorbox}

\section{Conclusion \& Discussion}
\label{sec:discussion}
In this work, we aim to achieve the twin goals of improved robustness and generalization in   SSL for code, specifically in \cllower. 
We realize this by proposing two improvements--adversarial positive views in \spc\cllower, and a staggered \spc\advtr schedule during fine-tuning.
We find that each of these proposals provides substantial improvements in both the generalization and robustness of downstream models; their combination, \ourmodel, provides the best overall performance.
Given the growing adoption of SSL-based models for code-related tasks, we believe our work lays out a framework to gain a principled understanding into the working of these models. Future works should study this problem while attempting to also establish a theoretically sound foundation. 

When compared to SSL for vision, it seems   SSL for codes benefits from milder adversarial training during fine-tuning.
Stronger attack methods for code might be needed to explore this phenomenon further. 
It will also be beneficial to understand how code models respond to perturbations, and to contrast it to our understanding of continuous data perturbations   in vision.


\section*{{Acknowledgement}}
This work was partially funded by a grant by MIT Quest for Intelligence, and the MIT-IBM AI lab.
The work of J. Jia and S. Liu was also supported by 
 National Science Foundation (NSF) Grant IIS-2207052.

\bibliographystyle{ieeetran}
\bibliography{IEEEtran,bibs_SL/ref_SL_adv,bibs_SL/ref_SL_BLO, bibs_SL/ref_SL_fair_self}

\end{document}